\documentclass[a4paper,11pt]{article}
\usepackage{jinstpub} 
\usepackage{lineno}

\title{\boldmath Hydra: Computer Vision for Data Quality Monitoring}

\author[1]{Thomas Britton,}
\author{Torri Jeske,}
\author{David Lawrence,}
\author{Kishansingh Rajput}
\affiliation{Thomas Jefferson National Accelerator Facility,\\
Newport News, VA, USA}

\emailAdd{tbritton@jlab.org}

\abstract{Hydra is a system which utilizes computer vision to perform near real time data quality management, initially developed for Hall-D in 2019. Since then, it has been deployed across all experimental halls at Jefferson Lab, with the CLAS12 collaboration in Hall-B being the first outside of GlueX to fully utilize Hydra. The system comprises back end processes that manage the models, their inferences, and the data flow. The front-end components, accessible via web pages, allow detector experts and shift crews to view and interact with the system. This talk will give an overview of the Hydra system as well as highlight significant developments in Hydra's feature set, acute challenges with operating Hydra in all halls, and lessons learned along the way.}

\begin{document}
\maketitle
\flushbottom

\section{Introduction}
\label{sec:intro}
Physics quality data is expensive and time-consuming to obtain. This expense underlies the importance of having good data quality monitoring (DQM).  In GlueX, where Hydra was first developed and deployed, DQM consisted of an iterative series of checks and rechecks which takes data from its raw form, through calibration, to reconstruction, and finally when integrated into a physics analysis. This process is coordinated by the Online Monitoring Coordinator, the Offline Monitoring Coordinator, and the Physics Analysis Coordinator.  The first and arguably most critical step of this process is the initial data collection, because data lost to problems with detectors (e.g. the failure of any electrical components) are unrecoverable.  To protect the data acquisition in GlueX shifts of two people, responsible for managing the acquisition, are instructed to monitor Rootspy \cite{RootSpy}, a program which collects statistics from the current data taking run and displays a variety of plots to shift takers.  Every day the Online Monitoring Coordinator looks over hundreds of plots generated based on the previous day of running and writes a short monitoring brief.  This method, while useful for catching issues that might have gone unnoticed by shift crews, places more burden on people.

Hydra was developed to alleviate this burden and free people up to work on things humans are naturally better at (e.g. making data taking decisions).  Hydra aims to be an extensible framework for training and managing AI, which leverages the recent developments in computer vision, for near real-time monitoring.  It is comprised of a Python back-end, supported by a MySQL database, and a Web-based front-end which allows for users to interact with and monitor the Hydra system(s). To shorten the runway for production Hydra was built primarily with computer vision in mind.  This means that Hydra can, out of the box, be applied to a variety of image classification tasks.

\section{The Hydra Back-end}
\label{BE}

\begin{figure}
    \centering
    \includegraphics[width=0.6\linewidth]{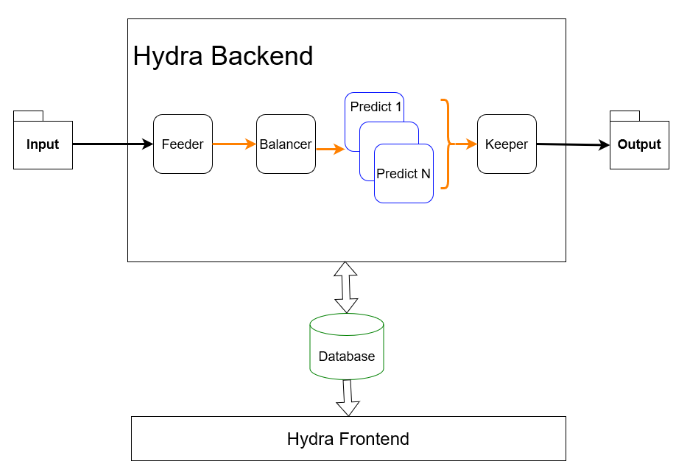}
    \caption{A simplified diagram of the Hydra system. The orange arrows indicate the path of inference orders through Hydra's back end processes. The MySQL database is indicated in green.}
    \label{fig:hydra-process}
\end{figure}

Hydra runs images through a modular, multi-step workflow comprised of a set of Python scripts all of which interact with an underlying MySQL database. An overview of the system is shown in Fig. \ref{fig:hydra-process}.  Python was chosen to more easily support the use of Tensorflow \cite{tensorflow2015-whitepaper} and Tensorflow-produced models, which formed the foundation for Google's Inception v3 \cite{szegedy2015rethinking}. Although InceptionV3 is the model topology most used in Hydra, it is not a requirement and users can implement their own model so long as it is compatible with Tensorflow.  The modular components of the Hydra Back-end are described in the following subsections.

\subsection{Database}
A comprehensive MySQL database supports both the back end and front end of Hydra. The database stores the location of the trained model files as well as the mapping that associates the various models to their corresponding plot type.  This allows Hydra to run inference on all of the various plot types using the corresponding model.  To support the front end,  the database stores references to all available images from all available plot types, all valid labels to every plot type (different plot types need not have the same classifications), permissions lists for labeling each plot type, the labels to every plot, model classification for every plot (when a model's performance is analyzed), the training set used to train every model, every in-situ inference performed by every model, and the last few minutes of the current run.  This coverage allows for extensive data mining, not only of the components of Hydra but also of overall data quality.  For example, by recording all of this information Hydra has spotted electronic components in the process of failing, flickering on and off at time scales far shorter than what shift takers could detect using the cumulative plots.  This depth of data tracking also allows for the pinpointing of the starting of detector problems to mere minutes (shorter if image rates are higher).

\subsection{Feeder}
Hydra's Feeder is a light-weight script that watches an input directory for new images that need to be analyzed and ensures they are of the right shape to be fed into the corresponding model. If necessary, the image is resized to match the input shape of the model as recorded in the database. As long as this is done consistently, artifacts produced as a result of resizing have not proven to be an issue.  Once the image is confirmed to be the right shape for model consumption Feeder sends an "Inference Order" to the load balancer via a ZeroMQ message.

\subsection{Load Balancing}
The Load Balancer is responsible for distributing the Inference Orders it receives to a configurable number of Predict processes.  It does this in Round Robin amongst the N Predict processes.  It does not modify the Inference Order in any meaningful way (it only adds its processing time to the message metadata for monitoring) and merely passes the message on.  Because of this simple pass through the Balancer takes about $10\mu s$ to process an Inference Order.

\subsection{Predict}
The Predict stage is where inference is run and a report is generated.  Each copy of Hydra Predict has its own buffer to store Inference Orders and then processes them on a First-in-First-out (FIFO) basis. The script uses the Inference Order to identify which image needs to be processed and internally maps that image to the specific model (called a Hydra Head internally) trained to process it.  Once processed a "Report" is created and transmitted via ZeroMQ to the Keeper process.  This report contains meta data about the image, its processing, and the model's labels and normalized output weights, from which the classification for the image can be determined.  In the event that a model classifies an image as "Bad" it automatically generates a Gradient weighted Class Activation Map (gradCAM) \cite{Selvaraju_2019} to be included in the submitted report. It is important to note that this stage does not do anything other than manage the running of Hydra's various models and generating a Report.  Actions (e.g. inference storage, RunTime displays, unbiased collection etc) are all taken by the Keeper, described in the next subsection. 

\subsection{Keeper}
The Keeper process takes various actions when it receives a Report from the Predict process(es).  It records the inference in a RunHistory table and records the image, the gradCAM heatmap (if applicable),  as well as a few key pieces of inference in a RunTime table.  Based on the model's classification and the collect percentage, Keeper makes a determination of what images to keep for labeling. Keeper prioritizes the collections of examples where the image is classified as "Bad" or the model's confidence is below threshold as configured in/read from the database on a per model basis. In addition, Keeper selects a random sampling, based on a configurable collect percentage, of images it sees. These images are used in future model training as well as in the monitoring of the model's in-situ performance. These examples provide very useful examples for future training to increase model robustness in problem classification and increase model confidence. 

\section{The Hydra Front-End}
Hydra has a web-based front end which acts as the primary interface with the Hydra system. Various pages have been developed such that users can label images, evaluate model performance, view near real time classifications of incoming data, inspect data from the previous day, and monitor the Hydra systems status.  A web based user interface enables users to perform data quality monitoring remotely further adding to the robustness of any DQM policies already in place.  The various web pages that comprise the Hydra Front-end are described below.

\subsection{Labeler}
\begin{figure}[htb]
    \centering
    \includegraphics[width=0.6\linewidth]{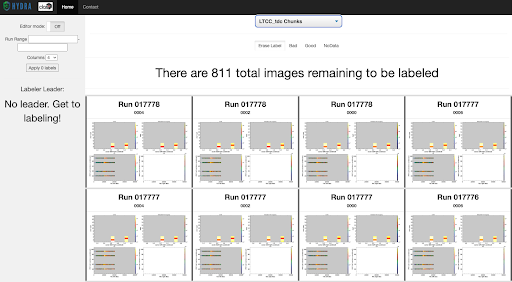}
    \caption{Snapshot of the online Labeler. Users can select the plot to label, select the appropriate label and label thousands of images quickly.}
    \label{fig:labeler}
\end{figure}

All supervised learning techniques require labeled training data, which can be costly to obtain. For this reason, a substantial amount of time was devoted to reduce the cost (e.g. expert's time) of labeling. The Labeler, shown in Fig. \ref{fig:labeler} is based on a palette system whereby users who have permissions can select a label, referenced by a custom color, from a palette of labels and "paints" a set of images.  To aid in the efficiency of labeling, shortcuts are provided to label blocks of images with the same label at the same time.  After this the user may "Apply" the labels which stores the labels in the database and removes those images from the set of unlabeled images and replaces them with new to-be-labeled images (as applicable). An Editor mode is provided where users can change the label of a previously labeled image.  In this mode labeled images can be filtered by both time and given label, allowing users to quickly filter through potentially hundreds of thousands of images to make the needed corrections.  With the labeling web page experts are able to label images at a rate of about ten thousand images per hour.

\subsection{Model Training Report Page}
Training is an iterative process; after training a new model an interactive report is produced which enumerates the differences between the model's classification and the human's classification.  It is through analysis of these differences that potential training problems can be identified and corrected. Commonly, human errors in labeling (usually a few percent) can be found and corrected.

\subsection{Library}
\begin{figure}[htb]
    \centering
    \includegraphics[width=0.6\linewidth]{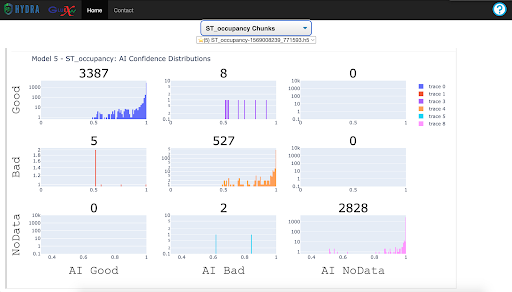}
    \caption{The library page displaying the enhanced confusion matrix for the active model (indicated with the star icon) is shown.}
    \label{fig:library}
\end{figure}

The Library page, shown in Fig. \ref{fig:library}, provides information regarding training for each of Hydra's models. It includes the sampling method used when generating the training set, training set size, and an enhanced confusion matrix. Active models, those used in production, are also indicated. The Enhanced Confusion Matrix (ECM) is the standard confusion matrix with a plot, per matrix cell, of the output weight of the model for each classification. With the ECM, a per label confirmation threshold can be applied. This threshold is a real valued number between 0 and 1 and acts to combat both false positives and false negatives. A model's classification is considered "confirmed" if and only if the output weight for the classification falls above the threshold. Each threshold can be modified, with valid permissions, via the web page. Default threshold settings are determined by the maximum effective F1 score in order to minimize false positives and false negatives. An image with an "unconfirmed" classification will be flagged by Keeper for labeling.

\subsection{Status}
The Status Page gives a near real time view of the computational health of Hydra.  It does this by displaying histograms of the time each of the back end processes took to process individual images, or inference orders, over the last 24 hours. A scatter plot displays the average processing time for each back end process, where the average is taken over the set of images, as a function of the experiment's run number.  Deviations in the individual distributions or average could indicate a technical fault with Hydra that includes, but is not limited to, a communication issue with the database server or issues with the underlying file system.  These issues can inhibit Hydra and are, for the most part, easily corrected once identified.

\subsection{Run}
\begin{figure}[htb]
    \centering
    \includegraphics[width=0.6\linewidth]{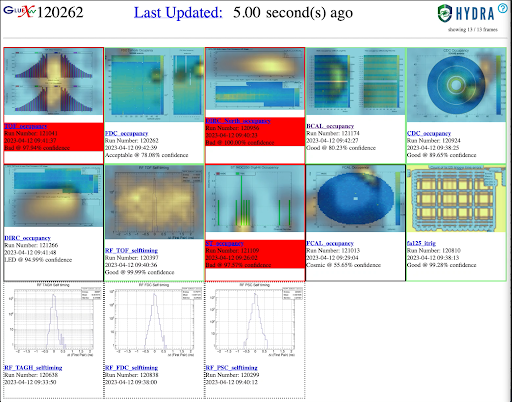}
    \caption{The Hydra Run page displaying near real time classifications along with gradCAM heatmaps overlayed on the images. These help indicate problematic areas of the images to aid the shift crew in diagnosis.}
    \label{fig:run}
\end{figure}

The Hydra Run page, shown in Fig. \ref{fig:run}, is up in the various Counting Houses Hydra is deployed in.  It provides a security camera style view of the images Hydra analyzes with Hydra's sentiment in real time. The page auto-updates, styling the various frames according to the classification Hydra gives the image.  Confirmed Bad and unconfirmed images are moved to the top left of the image grid and given visual indications of the model's classification with an optional setting to display the gradCAM heatmap overlayed on the image. Custom alarms can be added depending on the preferences of each experimental hall. All of this helps guide shift takers to potential issues in an easy to digest, interpretable way.  The Hydra run page also provide links to other Hydra web pages which may be of use to shift takers.

\subsection{Log}
The Hydra Log page provides an overview of all confirmed bad and any unconfirmed plots from the previous 24 hours. This page is often used as a daily brief during run coordination meetings to provide a quick snapshot of the prior day's data quality. Any problems that Hydra observed in the previous day can be identified and discussed with the relevant detector experts. 

\begin{figure}[htb]
    \centering
    \includegraphics[width=0.6\linewidth]{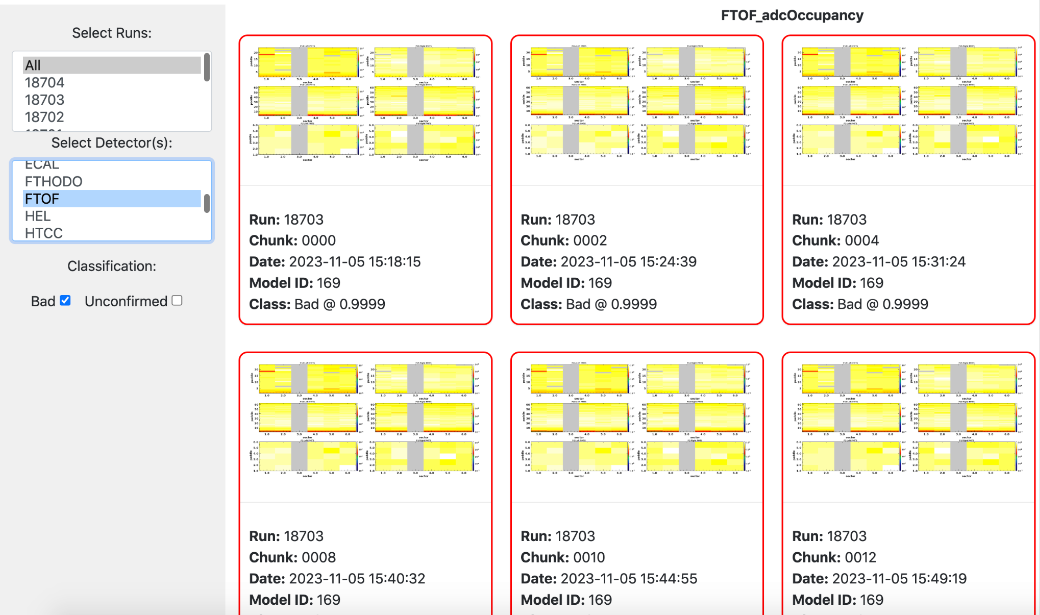}
    \caption{The Hydra Log page displaying confirmed bad plots from the previous day. }
    \label{fig:log}
\end{figure}

\subsection{Grafana}
Each inference from Hydra is stored in the database.  There are many different ways to visualize and interpret this data.  While not a required part of Hydra, the deployments at Jefferson Laboratory (JLab) are supplied with a Grafana \cite{Web:Grafana:Docs} server to visualize all of the inferences of Hydra.  To accomplish this the inferences are viewed as a set of time series data, one set per plot type. Because every label and every output weight for those labels are recorded it is possible to query any time window and display the series of output weights for all labels in a scatter plot.  Anecdotally, there have been occasions where the the growth in Bad output weights and corresponding drop of Good output weights, even in a non-monotonic or sinusoidal manner, precedes certain detector failures.  This behavior also signals that the model might need to be retrained with newly labeled images. While no conclusive robust study has been performed it is conceivable that this data may prove to have some predictive power when it comes to certain data acquisition problems.

\section{Deployments}
Currently, Hydra is deployed in all of JLab's experimental halls, with Halls B and D seeing the most active use. When Hall-D was taking data Hydra was in its infancy and even still discovered many problems, some even subtle enough that they were missed or would have been missed by detector experts. After deployment in Hall-B,  Hydra quickly set to work identifying many issues with super-human performance.

Deploying Hydra to all experimental halls came with a fair share of challenges.  The first being making a system that can easily integrate with the existing, but distinct, monitoring systems already in place in each hall.  To overcome this Hydra employs hall-agnostic technologies, such as using images in the place of underlying histograms values obtained from ROOT \cite{rene_brun_2020_3895860} trees where the layout of the trees would need to be known a priori. These images are trivially derived from ROOT based monitoring systems, allowing for quick deployments of the Hydra system.  As Hydra does not seek to replace human based monitoring, it is deployed in as parasitic a way as possible.  It does this by requiring only a standard naming convention for image files, an input directory, and a few directories for short and long term storage of images Hydra has selected to keep.  

\section{Developments}
The goal of Hydra is to build a self sustaining system in as many ways as possible.  It should intelligently suggest training, it should error correct whenever possible, and should be transparent and interpretable to humans. To accomplish this much of Hydra's road map for development can be broken down into coming from one of three main tracks:
\begin{enumerate}
    \item Features to expand on Hydra's capabilities to detect and diagnose issues.
    \item Features to enhance human control over the systems of Hydra and their operation.
    \item developments to make Hydra more computationally efficient.
\end{enumerate}

In order to expand Hydra's capabilities related to detection and diagnosis, a multistage analysis pipeline is being developed. The first step of this pipeline includes generic anomaly detection, such as implementing a Siamese model \cite{koch2015siamese}, before further diagnostics. Additionally, the ability to mask regions of images that should be ignored would allow on Hydra to focus on important data and avoid red herrings. Hydra's user interface is under continuous development to enhance user control and administration of the system without requiring knowledge of the back end processes or the MySQL database. Deployment to all experimental halls requires simultaneous monitoring of Hydra's health, database monitoring, and the current running conditions in each hall to ensure consistent, stable operations. An administrative interface is being developed to provide interpretable and actionable information regarding Hydra's health.


\section{Conclusion}
The Hydra system is a framework for managing the training and deployment of models for real time data quality monitoring.  It is deployed in all of the experimental halls at Jefferson Laboratory and successfully detects data quality issues at a super-human level.  It does this in a hall agnostic way with a robust web-based front end and back end, both of which are supported by a MySQL database.  Real-time views of incoming data are viewable from anywhere in the world, but most importantly from the counting houses where shift crews can get an at-a-glance look into Hydra's monitoring. Visual and, in some cases, audible alarms notify the shift crew when Hydra detects a problem. The visual indicators include a color determined by the classification and a gradCAM heatmap to aid the shift crew in determining the problematic regions of the image. Since Hydra looks at each image consistently and more frequently than the shift crews, corrective action(s) can be taken sooner. This is especially important while the shift crew is simultaneously responsible for the more complex tasks of running experiments in addition to data quality monitoring. Hydra is under very active development and looks to grow in robustness of detection, computational efficiency, and in human interface capabilities.

\acknowledgments

Jefferson Science Associates, LLC operated Thomas Jefferson National
Accelerator Facility for the United States Department of Energy under
U.S. DOE Contract No. DE-AC05-06OR23177. 
after the paper has been written.


 \bibliographystyle{JHEP}
 \bibliography{main.bib}


\end{document}